\pdfoutput=1

\documentclass[11pt]{article}

\usepackage[final]{acl}

\usepackage{times}
\newif\ifcomments
\commentsfalse

\usepackage{times}
\usepackage{latexsym}

\usepackage[T1]{fontenc}

\usepackage[utf8]{inputenc}

\usepackage{microtype}

\usepackage{adjustbox}
\usepackage{tabularx,rotating}
\usepackage{microtype}
\usepackage{graphicx}
\usepackage{subcaption}
\usepackage{graphicx}

\usepackage{longtable}
\usepackage{booktabs}
\usepackage{multicol}
\usepackage{multirow}
\usepackage{nicematrix}

\usepackage{hyperref}
\usepackage{tikz}

\usepackage[bibliography=common]{apxproof} 
\newtheoremrep{theorem}{Theorem}[section]

\usepackage{booktabs}
\usepackage{amsmath}
\usepackage{amssymb}
\usepackage{mathtools}
\usepackage{amsthm}
\usepackage{bbm}
\usepackage{soul}
\usepackage{algorithm}
\usepackage{algpseudocodex}
\algnewcommand\Inputs{\item[\textbf{Input:}]}%
\algnewcommand\Outputs{\item[\textbf{Output:}]}%
\algnewcommand\Params{\item[\textbf{Params:}]}%
\usepackage{xcolor}

\usepackage{nomencl}
\makenomenclature
\renewcommand{\nomname}{}

\makeatletter
\patchcmd{\thenomenclature}{%
  \section*{\nomname}
}{%
}{}{}
\makeatother

\DeclareMathOperator*{\argmax}{arg\,max}

\usepackage[capitalize,noabbrev]{cleveref}

\commentsfalse

\ifcomments
\newcommand{\authorcomment}[2]{\tikz[baseline=(X.base)]\node [draw=#1,fill=#1!40,semithick,rectangle,inner sep=2pt, rounded corners=3pt] (X) {#2};}

\newcommand{\cz}[1]{\authorcomment{blue}{Chaoyi:} \textcolor{purple}{\textit{#1}}}
\newcommand{\lc}[1]{\authorcomment{red}{Lydia:} \textcolor{red}{\textit{#1}}}
\newcommand{\py}[1]{\authorcomment{orange}{Pinyu:} \textcolor{blue}{\textit{#1}}}
\newcommand{\je}[1]{\authorcomment{pink}{Jeroen:} \textcolor{pink}{\textit{#1}}}
\newcommand{\PY}[1]{{\color{blue}PY: #1}}
\newcommand{\PYB}[1]{{\color{blue}[PY: #1]}}
\else
\newcommand{\authorcomment}[2]{}

\newcommand{\cz}[1]{}
\newcommand{\lc}[1]{}
\newcommand{\py}[1]{}
\newcommand{\je}[1]{}
\newcommand{\PY}[1]{}
\newcommand{\PYB}[1]{}
\fi

\newif\ifbluecolor
\bluecolorfalse
\newcommand{\cmr}[1]{%
  \ifbluecolor
    \textcolor{blue}{#1}%
  \else
    #1%
  \fi
}

\theoremstyle{plain}

\theoremstyle{definition}

\theoremstyle{remark}

\usepackage{xspace}
\newcommand{\algo}{\texttt{Duwak}\xspace}

\title{\algo: Dual Watermarks in Large Language Models}

\author{Chaoyi Zhu \\
  TU Delft, \\
  Delft, Netherlands \\
  \texttt{c.zhu-2@tudelft.nl} \\\And
  Jeroen Galjaard \\
  TU Delft, \\
  Delft, Netherlands \\
  \texttt{J.M.Galjaard@tudelft.nl} \\\AND
  Pin-Yu Chen \\
  IBM Research, \\
  New York, USA\\
  \texttt{pin-yu.chen@ibm.com} \\\And
  Lydia Y. Chen\\
    TU Delft, \\
  Delft, Netherlands \\
  \texttt{lydiaychen@ieee.org}
  }

\begin{document}
\maketitle


\begin{abstract}
As large language models (LLM) are increasingly used for text generation tasks, it is critical to audit their usages, govern their applications, and mitigate their potential harms.
Existing watermark techniques are shown effective in embedding \textit{single} human-imperceptible and machine-detectable patterns without significantly affecting generated text quality and semantics.
However, the efficiency in detecting watermarks, i.e., the minimum number of tokens required to assert detection with significance and robustness against post-editing, is still debatable. 
In this paper, we propose, \algo, to fundamentally enhance the efficiency and quality of watermarking by embedding dual secret patterns in both token probability distribution and sampling schemes.
To mitigate expression degradation caused by biasing toward specific tokens, we design a contrastive search to watermark the \emph{sampling scheme}, which minimizes the token repetition and enhances the diversity.
We theoretically explain the interdependency of the two watermarks within \algo. We evaluate \algo extensively on Llama2 \cmr{and 
Vicuna }under various post-editing attacks, against four state-of-the-art watermarking techniques and combinations of them.
Our results show that \algo marked text achieves the highest watermarked text quality at the lowest required token count for detection, up to 70\% tokens less than existing approaches, especially under post paraphrasing. Our code is available at \url{https://github.com/chaoyitud/Dual-Watermarks}. 
\end{abstract}

\section{Introduction} \label{sec:intro}
Large language models (LLMs) are widely adapted for natural language tasks, including copywriting~\cite{gpt}, machine-translation~\cite{DBLP:conf/icml/0006HB23:MachineTranslation}, questioning and answering~\cite{Tourvorn2023LLMA2}, and code generation~\cite{DBLP:journals/corr/abs-2308-12950:CodeLLM}.
While LLMs achieve remarkable and human-like performance, there are increasing risks of abusing LLM's~\cite{kuditipudi2023robust} to produce incorrect and adversarial content on social media and to commit fraud in academic rights.
Watermarking LLM content is one of the essential solutions to govern the LLM applications and guardrail their misuse and harm to the society, even requested by the governmental policies~\cite{veale2021demystifying}.
Much like physical watermarks, embedding watermark signals on LLM-generated text provides the means to trace content to their generator as well as the LLM models that constantly evolve.

\begin{figure*}[!ht]
\centering
    \includegraphics[width=0.85\textwidth,clip,trim=0cm 0.5cm 0cm 0cm]{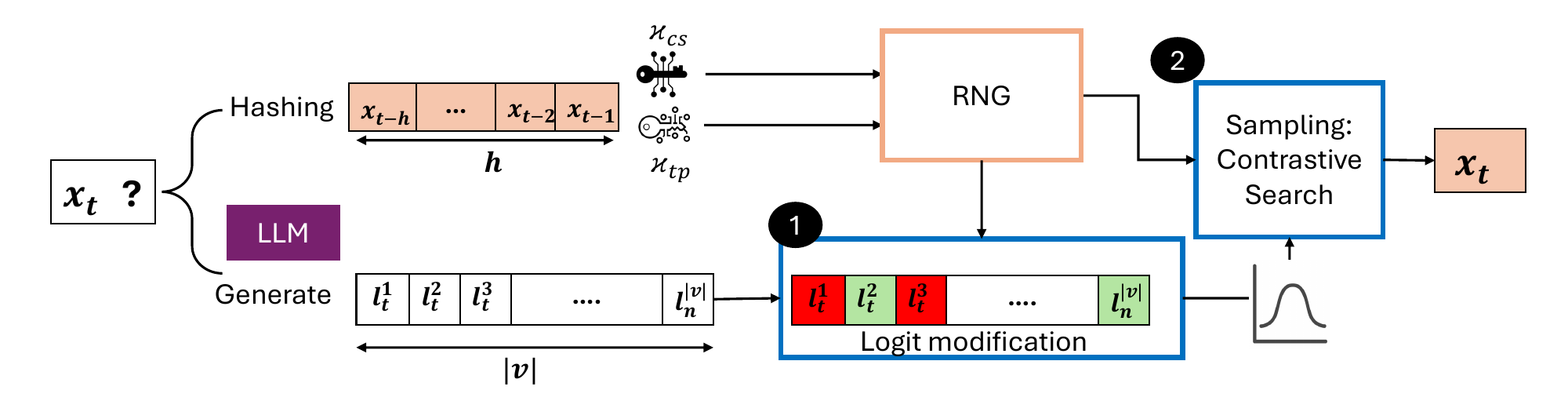}
    \caption{\algo: dual watermarking LLMs. To generate a token $x_t$, \algo embeds two secret patterns, governed by random number generation seeded by two private keys and prior tokens, via (i) pre-activation logit modification and (2) a contrastive search sampling strategy.}\label{fig:watermark}
\end{figure*}

Key criteria for watermarking generative language models are multiple folds: having minimal degradation of the generated content quality, imperceptible to humans for avoiding alteration, detectable by machines for rigorous auditing, and robust against post-text editing.
Recent studies show that a \textbf{single watermark} pattern can be hidden in generated text through either altering the underlying token probability distribution~\cite{kirchenbauer2023watermark,yoo2023multibit,fernandez2023bricksmutibit} or modifying the {sampling strategy}~\cite{kuditipudi2023robust,christ2023undetectable,OpenAI}. 
While the existing watermarks achieve multiple criteria, their practicability on short texts and post-edited text is limited, as the minimum number of tokens required for successful detection, e.g., low false positive rate, under those scenarios is high. 

In this paper, we propose a dual watermarking approach, \algo, which improves the watermark detection efficiency and text quality by embedding two independent secret patterns into the token probability distribution and sampling scheme.
To detect the watermark, \algo searches for the union of these two watermarks---the enabler for efficient detection with a low token count.
Under \algo, we first modify the pre-activation logits of pseudo-randomly selected tokens seeded by a function (i.e., hash) of a prior token sequence and a secret key, 
similar to green-red list  watermarking approaches~\cite{kirchenbauer2023watermark,yoo2023multibit, wang2023codable, zhao2023provable}.
Consecutively, we randomly split the token into the normal and watermark sampling set, which embeds an additional random number sequence seeded by a second secret key. 

The challenge lies in efficiently detecting watermarks without degrading quality. 
It is known that watermarking activation signals inevitably degrade the text quality due to the bias term on a pseudo-random selection of tokens~\cite{welleck2020ICLRUnlikelihood,kuditipudi2023robust}.
To counteract this degradation, we advocate the use of a quality aware sampling scheme---the contrastive search, which limits token sampling to top-k tokens resulting in the lowest similarity w.r.t. previous generated tokens. 
Unlike the popular cryptographic sampling, the contrastive search marks sampling patterns, thereby improving the text expression, improving the diversity of token selection and thus the watermarked text quality~\cite{Ren2023Semantic,Liu2024SemanticRobust}. 

Our contributions are summarized in the following:

\noindent\textbf{Improving watermark efficiency}, through \algo's joint dual watermarks patterns in the token probability distribution and sampling scheme.\newline
\noindent\textbf{Increasing generation diversity and robustness}, by avoiding expression degradation of watermarked text. \algo includes a novel quality-aware sampling scheme based on the contrastive search.\newline

\noindent\textbf{Empirical evaluation} showing the effectiveness of \algo against existing watermark solutions under nine post-editing attacks.
Thereby showing the minimum number of tokens required to reach detection accuracy is up to 70\% lower than related work, with nearly the best text quality and diversity. 


\section{Background}

\textbf{LLM text synthesis}
Large language models are typically transformer-based neural networks, denoted by $M$ and parameterized by $\theta$.
Internally, these models tokenize the vocabulary into a set, $\mathcal{V}$, and generate a token sequence indexed by $i$, $x_{i\ge 0}$, based on the prompt text, which is represented as a token sequence with negative index $x_{i<0}$.
Generally, generative LLMs `complete' a provided sequence (prompt) in an auto-regressive fashion, i.e., the token of $t$-th position is based on the prompt and thus far generated tokens, i.e., tokens $x_{i<t}$, from here on notated simplified as $x_{<t}$. 
The token generation consists of two stages. 
First, the LLM estimates the probability scores of the succeeding token $x^n_{t}$ for all $|\mathcal{V}|$ tokens at the position $t$, $\forall n \in \mathcal{V}$ by softmaxing the model's output logits, $l^n_t=l_\theta\left(\cdot \mid x_{<t}\right)^{n}$,
$$
p_{\theta}(\cdot|x_{<t})^{n}= \texttt{softmax} ({l_{t}})^{n}, \forall n\in \mathcal{V}. 
$$


The second step is to sample the token based on the estimated probability distribution.
Common sampling schemes differ in their objectives and complexity: greedy search, beam-search, top-k sampling~\cite{Fan2018TopkSearch}, nucleus-sampling (top-p)~\cite{HoltzmanICLR20topPSearch}, multinomial (random) sampling, and contrastive search~\cite{Yu2022NIPScontrastive}. 

\textbf{Watermarking LLM}
Watermarks are typically embedded in the process of next-token generation through altering: (i) the logit and probability ~\cite{kirchenbauer2023watermark,yoo2023multibit,Lee2023entropymark} and (ii) the sampling scheme~\cite{OpenAI,christ2023undetectable, kuditipudi2023robust}.
To change the probability distribution, the key idea is to split the vocabulary tokens into a green (preferred) and red list, $\mathcal{V}\in \mathcal{G}\cup  \mathcal{R}$, via a random number that is hashed from a secret key and an aggregate of previous $h$ tokens.
The number of green tokens is controlled by hyper-parameter $\gamma$ by taking $|\mathcal{G}| = \gamma |\mathcal{V}|$.
The logit values of green tokens receive a bias $\delta$, thereby increasing their probability estimates, thus increasing the likelihood of them being selected. 
The sampling scheme can remain the same as the original LLM.
Consequently, watermarked text is expected to have an increase in the number of green tokens. 
In contrast, sampling-based approaches are deterministic while keeping the model's next token probability estimate untouched.
\citet{OpenAI} propose an exponential scheme and choose the token $x_t = {\argmax}_{{n \in \mathcal{V}}}\left\{ (r^n)^{\frac{1}{p^n}} \right\}$, where $p$ is the unaltered probability vector and $r\in [0,1]^{\mathcal{|V|}}$, is the random number vector generated by hashing the prior $h$ tokens and the secret key.

\textbf{Detecting Watermarking}
Detecting watermarks requires inspecting a sequence of $N$ tokens and computing their watermark likelihood score, $S_N$.
The exact score computation depends on watermarking methods. In the case of logit modification through the green-red list~\cite{kirchenbauer2023watermark}, every token is classified into the green or red list based on the random split, conditioned on the random number sequence seeded by prior tokens and secret key.
The total number of green tokens is the score. 
As for the sampling approach, e.g., \citet{OpenAI}, computes a pre-determined threshold is exceeded by negative summation of $\sum_{i\in N} \ln( 1 - r_{i})$. 
Here the intuition lies in the fact that a token with low $p_i$ would require an $r_i$ arbitrarily close to $1$, thus limiting their contribution to the computed score.
This metric essentially measures the aggregate deviation from the expected distribution of tokens under the manipulation of random number vector $r$.

\textbf{Watermarking Measures}
There are multiple measures for watermarking algorithms: text quality, detection efficiency, and robustness. In terms of quality, perplexity~\cite{kirchenbauer2023watermark, DBLP:journals/corr/abs-2307-15992, kuditipudi2023robust} metrics, 
rating from another (larger) LLM~\cite{DBLP:conf/eamt/KocmiF23, piet2023mark}, and diversity~\cite{Yu2022NIPScontrastive, kirchenbauer2023reliability} are used to assess the (watermarked) LLM text. 
As for detection efficiency and robustness, it measures the number of $N$ tokens needed to achieve significant detection tests under different
 attacks, e.g., insertion, deletion, and paraphrasing~\cite{piet2023mark}.  

Z-statistic and p-value~\cite{kirchenbauer2023watermark} are commonly used to evaluate the significance of the detection test, assuming the detection scores follow the normal distribution with a mean of $\mu$ and standard deviation of $\sigma$. The null hypothesis of the detection test is that $H_0$: the text is unwatermarked. The Z-statistics represents the normalized observed score value, which is subtracted by the estimated mean and standard deviation. And, its corresponding p-value represents the probability of having a normalized score higher than observed $Z$ under the $H_0$, i.e., the text is not watermarked.

\section{\algo: Dual Watermarking}\label{sec:}
The objective of \algo is to maintain the watermarked text quality while keeping high detection efficiency, i.e., high detection confidence by inspecting a low number of tokens.
 \algo embeds two secret watermark signals sequentially in the token probability distribution and token sampling scheme as shown in Fig.~\ref{fig:watermark}. 
To mitigate the text distortion caused by modifying the token probability, we design a contrastive search sampling scheme that increases the diversity via selecting tokens with low similarity among the top-$k$ ones. 
We elucidate the interdependency through the joint watermarking scheme of 
\algo, demonstrating that the two watermarks can be integrated efficiently with an efficiency guarantee.

\subsection{Token Probability Watermark}
To generate token $x_t$ from a given prompt and prior generated token sequence, \algo first alters the token probability distribution $p_{t}  \rightarrow \hat{p}_{t}$ by altering the logit values for a subset of $n\in \mathcal{V}$. Specifically, a secret key $\kappa_{tp}$ and the prior sequence of window $h$, i.e., $x_{t-h\le t \le t-1}$, are inputs to a pseudo-random number generator, $\texttt{RNG}$, for generating a fixed-length pseudo-random number sequence.
Consecutively, each random number is used to split the token into binary types, i.e., green v.s. red. 
Generally, the secret keys used during watermarking are only known to the owner.
Such a design guarantees that only the watermark owner can identify and decode the watermarked tokens, embedding a layer of security and specificity within the generated text.
Following~\cite{kirchenbauer2023watermark}, a bias term, $\delta$, is added to the logit of tokens on the favored list, termed green list, while keeping logits of non-biased tokens, coined red list, remains unchanged.
As the token probability distribution is computed as taking the softmax function on the logit, shown in Eq.~\ref{eqn:token_probability}, the token probability distribution is thus modified, risking text quality degradation.
The higher the $\delta$ value, the higher the distortion to the probability and thus higher the possibility
of degradation in text quality.
We note that \algo is compatible with any probability modification proposed in existing watermarking algorithms, and we, in practice, adopt the algorithms derived in~\cite{kirchenbauer2023reliability}. 
More specifically, defining $p_{t}^n$ as, 
\vspace{-1em}
\begin{equation}
 \hat{p}_{t}^n = \frac{\exp\left(l^n + \mathbbm{1}\big[n \in \mathcal{G}\big] \delta\right)}{\sum_{i \in V} \exp\left(l^i + \mathbbm{1}\big[i \in \mathcal{G}\big] \delta\right)} \label{eqn:token_probability}
\end{equation}
where $\mathbbm{1}[c]$ is $1$ when $c$ holds, otherwise 0. 

\subsection{Contrastive Search Watermark}
One of the known limitations of LLM is anisotropic representation---repetitive wording and degenerated expression~\cite{DBLP:conf/emnlp/Ethayarajh19, Yu2022NIPScontrastive, DBLP:journals/tmlr/SuC23}. 
To avoid such degradation, \cite{DBLP:journals/tmlr/SuC23} define a self-similarity measure of token $x_t$\ with respect to all other tokens in the vocabulary $\mathcal{V}$, i.e., $x_{j \in V \setminus \{i\}}$.
A higher value of self-similarity suggests a more isotropic representation space.
To address the isotropic degradation, the token is then sampled to maximize the summation of the weighted token probability and the penalty of self-similarity.

We adapt such a contrastive search principle into a watermark sampling scheme in a sliding window manner. This approach not only incorporates a distinctive sampling scheme but also significantly enhances the diversity of text generation. Effectively reducing token repetition and mitigating text degeneration, leading to more coherent and varied output. Here, token at position $t$, are split into two sets, (i) $\mathcal{C}$ with a probability $\eta$, subject to contrastive search sampling, and (ii) $\overline{\mathcal{C}}$ with a probability $1-\eta$, where standard multinomial sampling is applied. The segmentation into $\mathcal{C}$ and $\overline{\mathcal{C}}$ is facilitated by a pseudo-random number generator that leverages a hashing value of previous tokens and a watermark key, $\kappa_{\text{cs}}$.


Contrastive searching sampling aims to reduce the similarity to the prior $L$ token sequence.
For all the contrastive set, we limit the selection to the top-k\je{Do with fix k?} tokens, i.e., $V_t^{(k)}$, with the highest $k$th probability. 
The top-k sampling is designed to reduce the risk that unlikely tokens are sampled~\cite{Lewis:ACL:2018:top-k}, reducing the search space of contrastive search. 
We then choose a token, $v\in V_t^{(k)}$ that maximizes the weighted probability and minimizes self-similarity with respect to the prior $L$ tokens. 

We first define the similarity between $x_t$ and $x_{t-L\le j<t}$ as the cosine distance between their hidden state, 
\(s\left(h_{x_{t}}, h_{x_j}\right)=\cos(h_{x_i},h_{x_j}), \)
where $h_{x_i}$ and $h_{x_j}$ represent the last layer hidden states in the model of token $x_i$ and $x_j$ respectively, and $\cos$ is the cosine-similarity between embeddings. 
Extending it to the $L$ window, the self-similarity of $x_t$ is computed as the maximum value with respect to all $L$ prior tokens, $x_{t-L\le j<t}$, i.e., 
\(
s_L(x_t)=\max_{ t-L\leq j < t} \left\{s\left(h_{x_t},h_{x_{j}}\right) \right\}. 
\)

A sliding window \(L\) increases generation efficiency by limiting the similarity computation to $L$ preceding tokens.
Moreover, it increases robustness against attacks by limiting the context on which the watermark is conditioned.
The token is finally chosen by maximizing the weighted probability, $\hat{p^v_t}$ and similarity penalty, $\cdot s_L(x_t^v)$, where $\alpha$ is a hyper-parameter that balances the importance of the weighted probability of the token against its self-similarity penalty.

\begin{equation}\label{eq:contrastive_search}
x_t =\underset{v \in \mathcal{V}^{(k)}}{\arg \max } \left\{(1-\alpha)\cdot \hat{p^v_t} - \alpha \cdot s_L(x_t^v)\right\}
\end{equation}
\begin{algorithm}[ht]
\caption{\algo Token Generation.}\label{alg:duwak_generation}
\begin{algorithmic}[1]
\Inputs $\theta, \kappa_{\text{tp}}, \kappa_{\text{cs}}$
\Params $\texttt{RNG}, k, L, \texttt{Hash}, \eta, s_{L}$
\Outputs $x_t \in \mathcal{V}$

\Function{DuwakGenerate}{}
\State{$seed \gets \texttt{hash}(x_{<t})$ }
\State $r \gets \texttt{RNG}(seed,\kappa_{\text{cs}})$ 

\Procedure{TokenProbWatermark}{}

  \State $\mathcal{G} \gets \texttt{RNG}(seed, \kappa_{\text{tp}})$ 

\State Compute $\hat{p}_{t}^n$ as Eq.~\ref{eqn:token_probability}
\EndProcedure
\Procedure{CSWatermark}{}
\If{$r<\eta$}
     \State $\mathcal{V}^{(k)}_{t} \gets \text{top}_k(\hat{p}_{t})$
    \State Contrastive search as Eq.~\ref{eq:contrastive_search}
\Else
    \State $x_{t} \sim \text{Multinomial}(\hat{p}_{t})$
\EndIf
\EndProcedure
\State\Return $x_{t}$
\EndFunction
\end{algorithmic}
\end{algorithm}
\setlength{\textfloatsep}{1em}

\subsection{Detection in \algo}
To detect the watermarks within a text sequence \(x\) of length \(T\), we employ hypothesis testing to differentiate between the null-hypothesis \( \mathcal{H}_0 \): ``the text is generated naturally'' and the alternative hypothesis \( \mathcal{H}_1 \): ``the text is generated with \algo.'' 

\begin{algorithm}[!ht]
\caption{\algo Watermark Detection.}\label{alg:duwak_detection}


\begin{algorithmic}[1]
\Inputs $\theta, \kappa_{\text{tp}}, \kappa_{\text{cs}}$
\Params $\gamma,T, \eta, M, L$
\Outputs $\text{p-value} \in [0, 1)$
\Function{DuwakDetection}{}
\Procedure{Compute\_$P_{tp}$}{}
    \State $\phi_{tp}\gets\sum_{t=1}^T \mathbbm{1}\big[x_t \in \mathcal{G}_t\big]$
    \State $z_{tp}=\frac{\phi_{tp}-\gamma T}{\sqrt{T \gamma(1-\gamma)}}$
    \State $ P_{tp}=1-\Phi(z_{tp})$
\EndProcedure
\Procedure{Compute\_$P_{cs}$}{}
    \State $P_{cs} \gets 1$
    \For{$\kappa_{m} \in \{\kappa_m\ | \ \kappa_m \neq \kappa_{cs}\}_{i=m}^{M}$}
    \State $
    P_{cs} \gets P_{cs} + \mathbbm{1}\big[\phi_{cs}^{(\kappa_{m})} \geq \phi_{cs}^{(\kappa_{cs})} \big]
    $
    \EndFor
    \State $P_{cs} \gets \frac{1}{1 + M} P_{cs}$
\EndProcedure
\State $P \gets 1 - F_{\chi^2}(x, 4) \quad \text{where} \quad x = -2(\ln(P_{tp}) + \ln(P_{cs}))$

\State \Return $P$

\EndFunction
\end{algorithmic}
\end{algorithm}
\setlength{\textfloatsep}{2em}

Given the incorporation of two distinct watermarks, we treat the detection of each as two separate and independent tests. 
We first detect token probability and constrastive search watermark independently and compute their p-values, namely, $P_{tp}$ and $P_{cs}$, against the full hypothesis that the text is not altered by token probability (constrastive search) watermark. We then apply  Fisher's method~\cite{fisher1922interpretation} to that combining p-values from these two independent tests into a single statistic follows a chi-square (\( \chi^2 \)) distribution with \(d=4\) degrees of freedom:
\[
-2 (\ln(P_{tp}) + \ln(P_{cs})) \sim \chi^2(4).
\]
Furthermore, the resulting p-value $P$, derived from the chi-square distribution, is given as:
\[
P = 1 - F_{\chi^2}\left(-2 \left(\ln(P_{kgw}) + \ln(P_{cs})\right), 4\right),
\]
where \( F_{\chi^2}\) is the cumulative distribution function (cdf) for the chi-square distribution. This provides a unified statistical measure to assess the presence of watermarks in the text.

To compute the p-values for both watermarks, we resort to a concept of score, \(\phi\), which represents the discernible discrepancy between watermarked and non-watermarked texts. Higher the score, stronger the evidence of watermarked text. We explain how to derive the p-values from their detection scores. 

\emph{P-value of token probability watermark} ($P_{tp}$). We use the number of detected green-listed tokens of the $T$ token sequence as the score, i.e.,
\(\phi_{tp}=\sum_{t=1}^T \mathbbm{1} \left[x_{t} \in \mathcal{G}_{t}\right] \),
where $\mathcal{G}_{t}$ is generated from $\texttt{RNG}\left(\operatorname{\texttt{hash}}\left(x_{<t}\right), \kappa\right)$, which based on the watermark key and preceding tokens.
To assert its significance, we apply a Z-test on $z_{tp}=\frac{\phi_{tp}-\gamma T}{\sqrt{T \gamma(1-\gamma)}}$ and then compute the corresponding p-value, as $P_{tp}=1-\Phi(z_{tp})$, where $\Phi$ is the cumulative distribution function of normal distribution.


\emph{P-value of contrastive search watermark} ($P_{cs}$). As the score distribution in non-watermarked text is unknown, our proposed score for the contrastive search watermark is based on self-similarity difference between the contrastive set, $\mathcal{C}$ and non-contrastive set $\overline{\mathcal{C}}$, split by using the key $\kappa$. Intuitively, the score is higher when the correct key, $\kappa_{cs}$, is used to split the set, compared to using arbitrary keys. To assert the statistical significance in the score difference, we propose to compare the scores between using the known private key $\kappa_{cs}$ and other $M$ randomly chosen keys, $\kappa_{1\le m \le M}$. 

We first formally define these two sets as, $\mathcal{C}$ and $\overline{\mathcal{C}}$.
Following that we define the score of contrastive search watermark using any key $\kappa$ as
\begin{equation}\label{eq:phi_score_kappa}
\phi_{cs}^{(\kappa)} = -\left( \frac{\sum_{t \in \mathcal{C}} s_L(x_t)}{|\mathcal{C}|} - \frac{\sum_{t \in \overline{\mathcal{C}}} s_L(x_t)}{T - |\mathcal{C}|} \right).
\end{equation}

We then compute the score for the key, $\kappa_{cs}$ and $\kappa_m$, and count the number of times that the score of using $\kappa_m$ is higher than $\kappa_{cs}$, Finally, we approximate the p-value of contrastive search as, 
\[P_{cs} = \frac{1}{M+1}\left(1+\sum_{m=1}^M \mathbbm{1}\big[\phi_{cs}^{(\kappa_{tp})} \geq \phi_{cs}^{(\kappa_{cs})} \big]\right).\]

\subsection{Theoretical Analysis}

The following theorem shows that two watermarks do not influence each other.

\begin{theoremrep}[Green List Tokens using topk]
Given $\mathcal{X}=\{x_1,\ldots,x_T\}$ from an LLM with green list fraction $\gamma$, and token $n$'s adjusted probability at $t$ follows
Eq. \ref{eqn:token_probability}. Define \(V_t^{(k)}\) as the set of top-\(k\) tokens by \(\hat{p}_t^n\),with $x_t\sim \text{Uniform}(V_t^{(k)})$. Given $
\mathbb{E}|V_t^{k}|_{\mathcal{G}} \geq \nu$, then the expectation and variance of the count of green list tokens \(|x|_{\mathcal{G}}\) in \(\mathcal{X}\) are bounded as follows:
\[
\mathbb{E}|x|_{\mathcal{G}} \geq \frac{\nu}{k}T,\ 
\operatorname{Var}|x|_{\mathcal{G}} \leq T \cdot \nu \left(k - \nu \right) k^{-2}.
\]
\end{theoremrep}
\begin{proof}
\textbf{Expectation:} The expected number of green list tokens, $\mathbb{E}|x|_{\mathcal{G}}$, is calculated as the sum of expectations over all tokens being selected from the green list across all $T$ steps. Given that $x_t$ is uniformly chosen from the set of top-$k$ tokens $V_t^{(k)}$, the probability of choosing a green list token at any step $t$ is the fraction of green list tokens in $V_t^{(k)}$, which is $\frac{\left|V_t^k\right|_{\mathcal{G}}}{k}$. Therefore:
\[
\mathbb{E}|x|_{\mathcal{G}} = \sum_{t=1}^{T} \frac{\left|V_t^k\right|_{\mathcal{G}}}{k} = \frac{1}{k} \sum_{t=1}^{T} \mathbb{E}\left[\left|V_t^k\right|_{\mathcal{G}}\right].
\]
Given $\mathbb{E}\left|V_t^k\right|_{\mathcal{G}} \geq \nu$, it follows that:
\[
\mathbb{E}|x|_{\mathcal{G}} \geq \frac{\nu}{k}T.
\]

\textbf{Variance:} For the variance, considering the sum of independent but not identically distributed Bernoulli trials, each trial's success probability is the fraction of green list tokens at step $t$, $\frac{\left|V_t^k\right|_{\mathcal{G}}}{k}$. The variance of a Bernoulli variable with probability $p$ is $p(1-p)$. Therefore, the variance of the total count of green list tokens is:
\[
\operatorname{Var}|x|_{\mathcal{G}} = \sum_{t=1}^{T} \frac{\left|V_t^k\right|_{\mathcal{G}}}{k}\left(1-\frac{\left|V_t^k\right|_{\mathcal{G}}}{k}\right).
\]
This simplifies to:
\[
\operatorname{Var}|x|_{\mathcal{G}} = T \cdot \mathbb{E}\left[\frac{\left|V_t^k\right|_{\mathcal{G}}}{k}\left(1-\frac{\left|V_t^k\right|_{\mathcal{G}}}{k}\right)\right].
\]
By applying Jensen's Inequality, due to the concavity of the function $f(x) = x(1-x)$ for $x$ in [0,1], and given $\mathbb{E}\left|V_t^k\right|_{\mathcal{G}} \geq \nu$, we derive:
\[
\operatorname{Var}|x|_{\mathcal{G}} \leq T \cdot \frac{\nu}{k}\left(1-\frac{\nu}{k}\right).
\]

\end{proof}

\begin{toappendix}
\begin{theoremrep}[Green List Tokens in \algo]
Given $\mathcal{X}=\{x_1,\ldots,x_T\}$ from an LLM with green list fraction $\gamma$, and adjusted probability of token $n$ at $t$ being

\[
\hat{p}_t^n=\frac{\exp(l^n+\mathbbm{1}[n \in \mathcal{G}]\delta)}{\sum_{i \in V} \exp(l^i+\mathbbm{1}[i \in \mathcal{G}]\delta)},
\]
define $V_t^{(k)}$ as top-k tokens by $\hat{p}_t^n$, with $x_t\sim \text{Uniform}(V_t^{(k)})$. The expectation and variance of green list tokens, $|x|_{\mathcal{G}}$, are bounded by:
\[
\mathbb{E}|x|_{\mathcal{G}} \geq A T,
\]
\[
\operatorname{Var}|x|_{\mathcal{G}} \leq {A T (1-A) (k+T-1)}k^{-1},
\]
with \[A=\frac{\gamma \beta S^{\star}}{1+(\beta-1) \gamma}\]
under an approximation of top-k sampling to a stochastic sampling based on adjusted probability without replacement for $k$ times. So ($x_t$) from $V_t^{(k)}$ approximates a stochastic process without replacement, reflecting the dynamic adjustment of probabilities as tokens are selected.
\end{theoremrep}
\begin{proof}
Referencing Lemma E.1 from \citet{kirchenbauer2023watermark}, when a token index $v$ is sampled from the watermarked distribution, the probability that the token is from the green list in the top-$k$ candidates is
\[
\mathbb{P}[v \in \mathcal{G}] \geq \frac{\gamma \beta}{1+(\beta-1) \gamma} S_{t}.
\]
For simplification in the proof, we rewrite $\left|V_t^k\right|_{\mathcal{G}}$ as $V_t^k$ and $\left|x\right|_{\mathcal{G}}$ as $x$.

Then, the expectation of the number of green list tokens in the top-$k$ can be expressed as:
\[
\mathbb{E}_k[V_t^k] = k \cdot \frac{\gamma \beta}{1 + (\beta-1) \gamma} S^t.
\]

Given $V_t^k$, the expectation of the number of green list tokens in $\mathcal{X}$ is
\[
\mathbb{E}_t[x \mid V_t^k] = \frac{T}{k} \mathbb{E}_k[V_t^k].
\]

Therefore, the total expectation of green list tokens in $\mathcal{X}$ becomes
\[
\mathbb{E}_t[x] = \mathbb{E}_t\left[\frac{T}{k} \mathbb{E}_k[V_t^k]\right] = \frac{T}{k} \cdot k \cdot \frac{\gamma \beta}{1 + (\beta-1) \gamma} \mathbb{E}_tS^t \geq T \cdot \frac{\gamma \beta S^{\star}}{1 + (\beta-1) \gamma},
\]

i.e., \[\mathbb{E}_t[X_G] \geq T A.\]

Then, the variance of the green list tokens in the top-$k$, considering a Bernoulli distribution:
\[
\operatorname{Var} V_t^k = k \frac{\gamma \beta S_t}{1+(\beta-1) \gamma}\left(1-\frac{\gamma \beta S_t}{1+(\beta-1) \gamma}\right).
\]
Consider the variance within the top-$k$ candidates, the variance of the green list token number in the whole sequence is
\[
\operatorname{Var} x =  \mathbb{E}_t[\operatorname{Var}[x \mid V_t^k]] + \operatorname{Var}[\mathbb{E}_k[x \mid V_t^k]].
\]
\[
\operatorname{Var}[x \mid V_t^k] = T\cdot \frac{V_t^k}{k}(1- \frac{V_t^k}{k}),
\]
\[
\mathbb{E}_t[\operatorname{Var}[x \mid V_t^k]]=T\left(\frac{\mathbb{E}_t[V_t^k]}{k}-\frac{\mathbb{E}_t[(V_t^k)^2]}{k^2}\right),
\]
given $\mathbb{E}_t[(V_t^k)^2]=(\mathbb{E}_t[V_t^k])^2+\operatorname{Var}_t V_t^k$ and $\mathbb{E}_t[V_t^k] \geq kA$, $\operatorname{Var}_t V_t^k\leq kA(1-A)$,
\[
\operatorname{Var}[\mathbb{E}_k[x \mid V_t^k]]=T^2 \cdot \frac{\operatorname{Var} V_t^k}{k^2}\leq T^2 \cdot \frac{kA(1-A)}{k^2},
\]
combine all terms together:
\[
\operatorname{Var} x \leq A T(1-A)(k+T-1) k^{-1}.
\]
\end{proof}
\end{toappendix}
In our theorem, we describe a bound that elucidates the interdependency between two watermarks. We model the contrastive search as akin to uniformly sampling from the top-\(k\) candidates. By influencing the selection among the top-\(k\) tokens based on historical similarity, with a large vocabulary size \(|\mathcal{V}|\) and a small \(k\), the process effectively approximates random selection. 
The theorem sets bounds on the expectation and variance of "green list" tokens, based on the limit of mean green token selection within the top-k candidates. This effectively outlines the interdependency between the two watermarks in our \algo.

\section{Evaluation}\label{sec:evaluation}
In this section, we first detail the evaluation setup on the LLM prompts and evaluation tasks.
The evaluation metrics are the quality of watermarked text and the token count needed to achieve certain detection $p$ values under normal conditions and various post-editing attacks.
We compare \algo against existing single watermark techniques and combinations thereof.

\subsection{Evaluation setup}
\lc{As we are short of space, lets cut 4.1 by at least half}

\textbf{Prompt}.
For evaluation, we use open-ended generation~\cite{DBLP:journals/tmlr/SuC23} and MarkMyWords'~\cite{piet2023mark} structured tasks. The detailed settings can be found in Appendix~\ref{app:prompt}.

\cmr{
\textbf{Models}.
In our experiments, we utilize two primary models: Llama2-7b~\cite{touvron2023llama} and Vicuna-7b-v1.5~\cite{zheng2024judging}.
}

\textbf{Evaluation metrics}.
To evaluate watermark methods, we use the following metrics: Diversity, MAUVE, Rating, and Detection efficiency.
Details on these metrics and their configurations are provided in Appendix~\ref{app:metrics}.

\textbf{Baseline}.
A summarized baseline overview is given in Tab.~\ref{tab:baseline} (i) the Kirchenbauer-Geiping-Wen (KGW) algorithm~\cite{kirchenbauer2023watermark}, Exponential (EXP)~\cite{OpenAI}, Binary~\cite{christ2023undetectable} (BINARY), Inverse Transform Sampling~\cite{kuditipudi2023robust} (ITS) and Contrastive Search (CS) (ours) are the single watermarking algorithm, and (ii) KGW-EXP, CS-EXP, and \algo (ours) are the dual watermark algorithms. 
We highlight where the watermark signals are inserted in the token probability or sampling. For dual watermarking schemes, we conduct the $\chi^2$ test on the p-value of each watermark as \algo. 

\textbf{Hyper-parameter setting}.
For a fair comparison across algorithms, we limit the hashing input to the first preceding token to generate watermark seeds for all watermarking algorithms.\je{Is this common practise?}\cz{not common, but here we fix the hash method for a fair comparison. ITS uses a fixed codebook as hashing input in~\cite{kuditipudi2023robust}}
As for the fraction of green tokens, $\gamma |\mathcal{V}|$ under KGW probability modification, we use a fixed $\gamma=0.5$.
The detection window of \algo is set as $L=50$ token, and the probability of contrastive search is $\eta=0.5$.

\begin{table}[!ht]
\centering
\resizebox{\linewidth}{!}{
    \begin{scriptsize}
        \begin{NiceTabular}{@{}lccccccc@{}}
        \toprule
        \diagbox{Comp.}{Alg.} & \textbf{KGW} & \textbf{EXP} & \textbf{ITS} & \textbf{BINARY} & \textbf{KGW-EXP} & \textbf{EXP-CS} & \textbf{\algo} \\ \midrule
        $\Delta P(x_{t}|x_{<t})$ & KGW & -   & - & -  & KGW  & - & KGW  \\ \midrule
        Sampler   & Multi & Exp     & Inverse  & Binary  & Exp   & CS & CS\\ \bottomrule
        \end{NiceTabular}
    \end{scriptsize}
}
\caption{Watermarking algorithms: token probability modification, and sampling scheme. `-' denotes no token probability distribution modification.}\label{tab:baseline}
\end{table}

\begin{table*}[!ht]
\centering
\resizebox{1\textwidth}{!}{ 
\begin{NiceTabular}{lcc|cccccccc}\toprule
Watermark     & Human & No Watermark & KGW & EXP & BINARY & ITS & CS & KGW-EXP & EXP-CS & \algo \\ 
\cline{1-11}
Diversity (\%) (\(\uparrow\))  & 93.62 & 86.66 & 81.41 & 39.58 & 44.56 & 78.72 & \textbf{86.53} & 17.90 & 83.83 & \ul{83.98} \\
MAUVE (\%) (\(\uparrow\))      & 100.0   & 82.36 & 75.5  & 55.87 & 55.57 & 79.02 & \ul{80.71} & 27.03 & 77.58 & \textbf{82.18} \\
Rating (\%) (\(\uparrow\))     & -     & 87.28 & 86.15 & 82.56 & \textbf{87.10}  & 86.25 & 83.74 & 77.14 & 83.91 & \ul{86.51} \\
Dection efficiency (\(\downarrow\)) & -  & -     & 113   & \ul{89.5}    & 847  & >1024 & >1024 & \textbf{79.5} & 572   & 94.5 \\
\bottomrule
\end{NiceTabular}
}
\caption{Comparison of watermarking methods on different metrics on Llama2-7b. Arrows point to the direction of better performance: a downward arrow (\(\downarrow\)) means lower is better, and an upward arrow (\(\uparrow\)) means higher is better. \textbf{Bold}/\underline{underlined} text means the best/second-best score.}\label{tab:watermark_comparison}
\end{table*}

\begin{table}[!ht]
\centering
\resizebox{0.5\textwidth}{!}{ 
\begin{NiceTabular}{lc|ccccc}\toprule
Watermark & No Watermark & KGW & EXP & BINARY & ITS & \algo \\ \midrule
Rating (\%) (\(\uparrow\)) & 84.1 & 82.1 & 82.0 & 82.2 & \textbf{83.4} & \ul{83.1} \\
Detection efficiency (\(\downarrow\)) & - & 101.5 & \textbf{71} & 252 & $>$1024 & \ul{82.5} \\
\bottomrule
\end{NiceTabular}

}
\caption{Comparison of watermarking methods on different metrics on Vicuna-7b-v1.5.}\label{tab:watermark_comparison_vicuna}
\end{table}

\begin{figure}[!ht]
    \centering
      \begin{subfigure}[b]{0.44\linewidth}
        \centering
        \includegraphics[width=\textwidth]{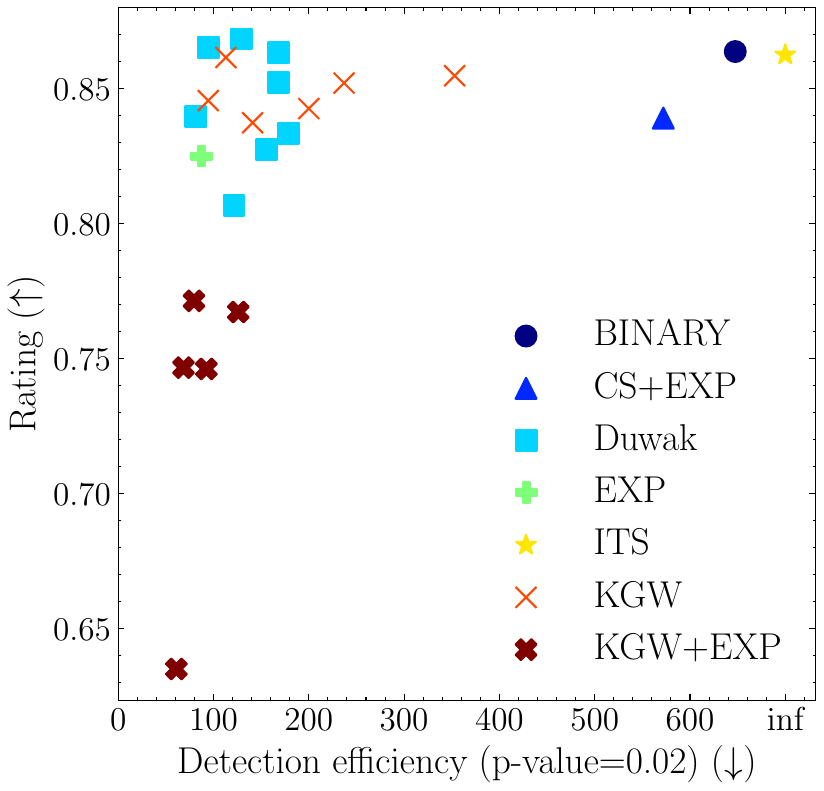}
        \caption{$p=0.02$}
        \label{fig:max_rating_0.02}
    \end{subfigure}
    \begin{subfigure}[b]{0.44\linewidth}
        \centering
        \includegraphics[width=\textwidth]{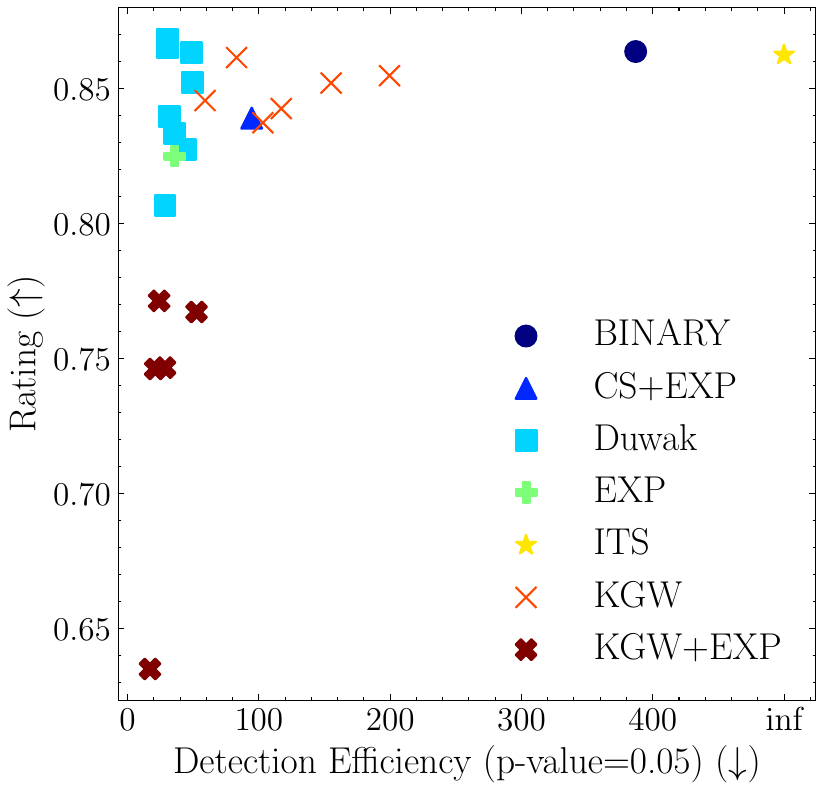}
        \caption{$p=0.05$}
        \label{fig:max_rating_0.05}
    \end{subfigure}

    \caption{Rating v.s. token efficiency under different watermarking methods and hyper-parameter settings for different detection $p$-values.}\label{fig:watermark_efficiency_front}
\end{figure}

\subsection{Results}
\textbf{Quality v.s. detection efficiency}.\label{subsec:quality_vs_efficiency} 
We summarize the overall results in Tab.~\ref{tab:watermark_comparison} and Tab.~\ref{tab:watermark_comparison_vicuna}, highlighting the difference among human, unaltered LLM, and watermarked LLM text from all the watermarking methods.
First of all, human-written text shows the highest diversity and MAUVE scores.
Regarding the quality of the watermarked text, \algo ranks as the first or the second-best method in terms of diversity, MAUVE, and rating, achieving similar results as the unaltered LLM text.
CS achieves the highest diversity and MAUVE as expected among the single watermarks.
Among dual watermarks, the direct combination of the common probability modification (KGW) and token sampling (EXP) 
deteriorates text quality due to the EXP sampling method, which heavily biases the modified token probability. 
Overall, including contrastive search improves the text quality to its CS-less counterpart.

The efficiency of detection of watermarks measures the number of tokens needed to detect watermarks with p-values of 0.02.
EXP-CS is the only exception because both watermarks are embedded in the sampling process and interfere with each other, arguing the risk of blending multiple watermarks.
On the other hand, a single watermark requires a significantly higher\je{Can we get a number here?} number of tokens, especially for BINARY, ITS, and CS, strengthen the watermarked text's robustness and quality.  

Fig.~\ref{fig:watermark_efficiency_front}\je{check reference!} provides a sensitivity perspective of watermark methods under different hyper-parameter settings and p-values, 0.02 and 0.05.
Specifically, different $\delta$ values are used in KGW probability modification.
\algo shows more consistent performance across all $\delta$'s compared to KGW, i.e., slightly higher rating and lower tokens with a lower variance. 
This trend continues for a p-value of 0.05, with a more pronounced difference in their required token counts\je{Add number!}. Specifically, when compared to the best KGW watermark, our algorithm requires $\sim40$ fewer tokens.
When p-values are smaller, the number of tokens needed for detection increases considerably\je{Add number!}.

\algo achieves the best quality efficiency ratio, high diversity, MAUVE, and rating, using fewer tokens to detect watermarks accurately compared to other watermarking methods.
\noindent\begin{table}[!hb]
\centering
\resizebox{0.95\linewidth}{!}{
\begin{tabular}{llccc}
\toprule
Attack                  & Conf.                & EXP           & KGW & \algo         \\ \midrule
\multicolumn{2}{l}{None}                & \textbf{89.5} & 113 & 94.5          \\
\multicolumn{2}{l}{Contraction}         & 88.5          & 114 & \textbf{87.5} \\
\multicolumn{2}{l}{Lowercase}           & \textbf{106}  & 146 & 113           \\
\multicolumn{2}{l}{Repetition\&deletion} & \textbf{83.5} & 108 & 87.0          \\

    {Paraphrase}                        & GPT3.5  & 238           & 322 & \textbf{193}  \\
{\multirow[t]{2}{*}{Misspelling}}   & 25\%    & 93.5          & 119 & \textbf{82.5} \\
                                    & 50\%    & 148           & 147 & \textbf{114 } \\
{\multirow[t]{2}{*}{Swap}}    & 5\%     & 83.0          & 113 & \textbf{77.5} \\
                              & 10\%    & 83.0          & 113 & \textbf{82.0} \\
{\multirow[t]{4}{*}{Synonym}}      & 25\%    & 90.5          & 118 & \textbf{81.0} \\
                              & 50\%    & 100           & 134 & \textbf{100}  \\
                              & 75\%    & 126           & 169 & \textbf{112}  \\
                              & 100\%   & 170           & 213 & \textbf{125}  \\
{\multirow[t]{2}{*}{Translation}}  & FR      & 118           & 147 & \textbf{114}  \\
                              & RU      & 156           & 195 & \textbf{148}  \\
{\multirow[t]{2}{*}{TypoAttack}}  & 5\%     & 221           & 221 & \textbf{177}  \\
                              & 10\%    & 389           & 337 & \textbf{301}  \\\bottomrule
    \end{tabular}
}
\caption{Attacked detection efficiency on Llama2-7b, lower is \textbf{better}.}\label{tab:attack_results}

\end{table}

\begin{table}[!hb]
\centering
\resizebox{0.95\linewidth}{!}{
\begin{tabular}{llccc}
\toprule
Attack                  & Conf.                & EXP           & KGW & \algo   \\ \midrule
\multicolumn{2}{l}{None}                & \textbf{71}   & 101.5 & 82.5          \\
\multicolumn{2}{l}{Contraction}         & \textbf{72.5} & 99    & 87.5          \\
\multicolumn{2}{l}{Lowercase}           & \textbf{108}  & 130   & 113.5         \\
\multicolumn{2}{l}{Repetition\&deletion} & \textbf{72.5} & 114   & 89            \\
{Paraphrase}                             & GPT3.5        & \>1024& 582   & \textbf{328}  \\
{\multirow[t]{2}{*}{Misspelling}}        & 25\%          & 124   & 128.5 & \textbf{116.5}\\
                                         & 50\%          & \textbf{82} & 96    & 86           \\
{\multirow[t]{2}{*}{Swap}}               & 5\%           & 84.5  & 96    & \textbf{84}   \\
                                         & 10\%          & \textbf{84} & 100.5 & 101.5        \\
{\multirow[t]{4}{*}{Synonym}}            & 25\%          & \textbf{80.5} & 118  & 91           \\
                                         & 50\%          & \textbf{97}   & 131  & 116          \\
                                         & 75\%          & 142.5 & 139.5 & \textbf{126}   \\
                                         & 100\%         & 206.5 & 156   & \textbf{126.5} \\
{\multirow[t]{2}{*}{Translation}}        & FR            & \textbf{102}  & 155  & 106.5        \\
                                         & RU            & \textbf{137.5}& 168  & 148          \\
{\multirow[t]{2}{*}{TypoAttack}}         & 5\%           & 212   & 209.5 & \textbf{185}   \\
                                         & 10\%          & \>1024 & \>1024& \textbf{316}   \\\bottomrule
\end{tabular}
}
\caption{Attacked detection efficiency on Vicuna-7b-v1.5, lower is \textbf{better}.}\label{tab:vicuna_attack_results}
\end{table}
\textbf{Post-editing attack robustness}.
Here, we evaluate the robustness of \algo under different post-editing attacks, i.e., attacks that alter the tokenization.
Specifically, we consider contraction, lowercase, misspelling, repetition, swap, synonym, translation, typo, and paraphrase attacks from MarkMyWords~\cite{piet2023mark}. 
Tab.~\ref{tab:attack_results} and Tab.~\ref{tab:vicuna_attack_results} present the efficiency of reaching a p-value of 0.02 under KGW, EXP, and \algo.
Such a selection is based on the observation in Tab.~\ref{tab:watermark_comparison} that only these three methods achieve reasonable text quality while inspecting roughly 100 tokens. 

In Tab.~\ref{tab:attack_results}, while EXP shows the best efficiency in the no-attack scenario (through significant inference quality), \algo requires significantly lower tokens for inspection in the presence of attacks, i.e., ranging between 6 to 70\%. 
The presence of attacks clearly increases the need to consider more tokens for all watermark methods.
Let's zoom into the performance of \algo against each of those attacks, in contrast to the cast of no attack.
TypoAttack significantly increases the detection difficulty and results in a more than $3\times$ increase in the number of tokens.
Misspelling and repetition\&deletion, swap, and synonym (25\%) are simple attacks, even reducing the number of inspection tokens. 
Paragraphs and TypoAttack are where \algo has the best performance, compared to EXP, the second-best policy. We attribute this difference to the two watermarks and no interference among them. Additionally, \algo benefits from incorporating two distinct watermarks that operate without mutual interference, thereby enhancing its robustness.
\cmr{In results from Vicuna-7b-v1.5, as shown in Tab.~\ref{tab:vicuna_attack_results}, we observe similar trends in performance. However, under some attacks, particularly simpler ones, EXP achieves better efficiency. Nevertheless, in more severe scenarios, especially with strong attacks like the paraphrase attack, \algo significantly outperforms EXP, demonstrating its robustness in handling more complex attacks.}

\begin{figure}[!ht]
    \centering

    \begin{subfigure}[t]{0.48\linewidth}  
        \centering 
        \includegraphics[width=\textwidth]{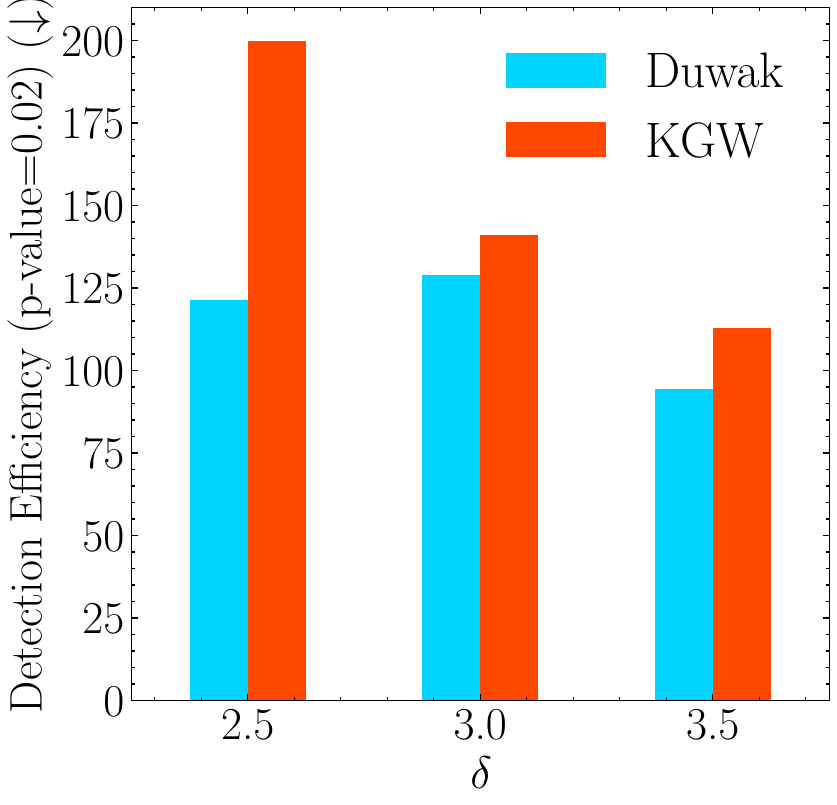} 
        \caption{No attack.}
        \label{fig:min_efficiency}
    \end{subfigure}
    ~
    \begin{subfigure}[t]{0.48\linewidth}  
        \centering 
        \includegraphics[width=\textwidth]{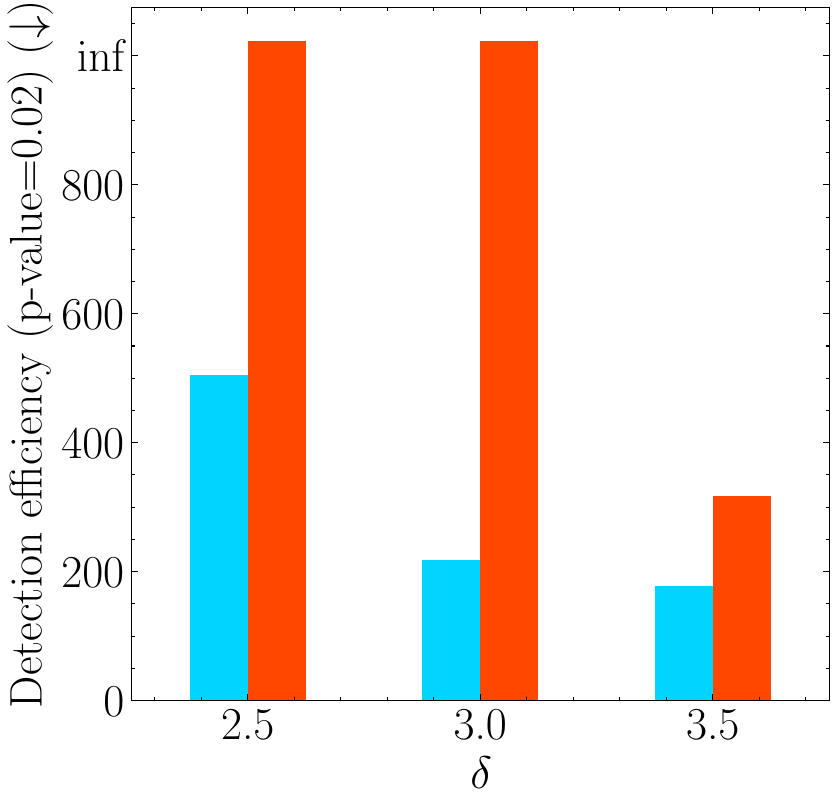} 
        \caption{Paraphrase attack.}
        \label{fig:min_efficiency_para}
    \end{subfigure}
    \caption{Detection efficiency ($\downarrow$) of \algo and KGW with equal hyper-config under varying $\delta$.
    }\label{fig:comp_cs_KGW}
\end{figure}

\textbf{Impact of contrastive search sampling}. 
Here, we highlight the impact of contrastive search compared to the single KGW watermark.
In Fig.~\ref{fig:comp_cs_KGW}, we show the rating and the number of inspected tokens to achieve a p-value of 0.02 under the different distribution shifting($\delta \in\{2.5, 3, 3.5 \}$) with clean and paraphrase attack versions. 
Unsurprisingly, \algo outperforms KGW due to the addition of contrastive search, such the advantage diminishes with increasing $\delta$.
When $\delta$ is large, e.g., 3.5, it introduces a large distortion in the generation probability, leaving little room for \algo to further improve the quality.
In the case of the challenging paraphrasing post-attacks, shown in Fig.~\ref{fig:min_efficiency_para}, one can observe the clear advantage of using contrastive search.
This observation again verifies our design of dual watermark, which is inherently more robust to the post-editing when compared to the token-level approaches of prior art.

\section{Related Studies} \label{sec:related}

Prior single watermark solutions embed the watermark signal at the token level with a modification of the generation process by modifying either the token probability distribution~\cite{lee2023SWEETwrote,wu2023dipmark,takezawa2023NSwatermark} or sampling scheme~\cite{OpenAI, christ2023undetectable, kuditipudi2023robust}.

\textbf{Watermark in token probability distribution.} \citet{kirchenbauer2023watermark} design the very first single-bit watermark method for LLM text generation, splitting tokens into a green and red list using a cryptographic key.

To further improve the text quality and robustness, subsequent studies modify the criteria of green-red splits. 
\citet{zhao2023provable} prove that global red-green splits improve robustness against post-editing attacks, whereas \citet{kirchenbauer2023reliability} propose to use the minimum hashed token to determine the red-green list. 

Furthermore, to improve the governance of watermarks and provide additional information, e.g., copyright and timestamp, multi-bit watermarks \cite{wang2023codable,yoo2023multibit,fernandez2023bricksmutibit} are proposed, introducing message-specific red-green lists. 
We note that such watermarks split the text into multiple sections, each of which has only a single watermark in their token probability, whereas our solution embeds up to two watermarks into a single token.

\textbf{Watermark in Sampling} 
Binary watermark~\cite{christ2023undetectable}
samples the token based on the comparison of the predicted probability and the pseudo-random presentation.
Because of the fixed length of pseudo-random numbers, the LLM can end up generating the same text for the same prompt.
\citet{kuditipudi2023robust} propose the usage of longer pseudo-random number sequences than the generated text itself and randomly choose the insertion location in the text to add the watermark.
\citet{hou2023sentencesampling} resort to watermarking via sentence-level sampling, which iteratively performs sentence-level rejection sampling until the sampled sentence falls within the watermarked region.

We note that orthogonal to watermark detection is more general detection of \textit{whether} text is synthesized by LLMs~\cite{Solaiman:CoRR:19:LogLikelihood, Gehrmann:ACL:19:GLTR,Mireshghallah:CoRR:23:DetectGPTBlackBox, Mitchell:ICML:23:DetectGPT,hu2023radar}. However, as traceability to specific models cannot be provided, these detection works are limited in their application for the governance of synthesized text.

\section{Conclusion}
In this paper, we propose a dual watermark scheme for LLM, \algo, which embeds human imperceptible and machine detectable watermarks in token probability distribution and sampling schemes.
Combining two watermarks significantly decreases the minimum number of tokens for detecting watermarks with a desirable false positive rate, especially when encountering post-editing attacks.
To avoid text quality degradation due to watermarking token probabilities, we design a contrastive search sampling scheme that samples tokens with the lowest similarity.
\lc{update the theory part}We show the effectiveness of \algo by providing a theoretical lower bound on the watermarked tokens and extensive empirical evaluation.
Compared against existing single watermark solutions and combinations thereof, \algo provides a better watermarked text quality.
This is especially highlighted in terms of diversity, and robustness against nine post-editing attacks, using up to 70\% less tokens for detection. 

\section{Limitation}
This study introduces advancements in watermarking techniques for Large Language Models (LLMs) through \algo, while also recognizing certain limitations that warrant future investigation. Firstly, our approach's effectiveness is contingent on the specific characteristics of the LLMs evaluated, primarily Llama2. Consequently, the applicability of \algo to different models and subsequent versions of LLMs is a subject that merits further exploration. Moreover, our evaluation was restricted to text-generation tasks. The extension of our methodology to encompass additional tasks, such as the generation of mathematical proofs or code, remains an area requiring in-depth study.
Additionally, \algo necessitates conducting two separate detection processes for each watermark, which results in a decrease in detection time efficiency compared to single watermark methods.

\section*{Impact Statements}
With the popularity of large language models and their applications, embedding watermarks into their generated content is an essential step toward trustworthy and responsible AI technology development and deployment. Our findings of improved watermark detection performance and utility provide novel insights into the research and practice of watermarking for large language models.

\section*{Acknowledgment}
This work has been partly funded by the Dutch National Science Foundation Perspectief Project, DEPMAT.

\bibliography{reference}

\appendix
\onecolumn
\section{Nomenclature}
\printnomenclature

\nomenclature{$\boldsymbol{\theta}$}{Large language model parameters used for text generation.}
\nomenclature{$\kappa_{\text{cs}}$, $\kappa_{\text{tp}}$}{Secret keys used for embedding watermarks in the text.}
\nomenclature{$\delta$}{Bias term added to the logits of tokens on the green list to alter their probabilities.}
\nomenclature{$\eta$}{Probability determining whether contrastive search or multinomial sampling is used for token generation.}
\nomenclature{$\alpha$}{Parameter balancing the importance of token probability and self-similarity in contrastive search.}
\nomenclature{$\gamma$}{Portion of the vocabulary designated as the green list in the token probability modification process.}

\nomenclature{$\hat{p}_{t}^n$}{The probability distribution over tokens after applying watermark modifications.}
\nomenclature{$l_{t}$}{The raw outputs of the LLM model for the next token, before applying the softmax function.}
\nomenclature{$\mathcal{G}$}{A subset of tokens selected for next word generation, influenced by a watermark key.}
\nomenclature{$s_L(x_t)$}{A measure of a token's similarity to its preceding tokens within a sliding window of length $L$.}

\nomenclature{\texttt{RNG}}{A function generating pseudo-random numbers based on a seed and possibly a key.}
\nomenclature{\texttt{Hash}}{A function used to generate a hash value based on the current context.}

\nomenclature{$k$}{Top-k parameter defining the number of top predictions considered in the generation process.}
\nomenclature{$L$}{Sliding window length used in contrastive search to compute token similarity.}
\nomenclature{$\mathcal{V}$}{The set of all possible tokens the LLM model can generate.}
\nomenclature{$H_{V}$}{The representation of tokens in the model's hidden layer.}

\nomenclature{$\phi_{tp}$}{A score computed for token token probability watermark}
\nomenclature{$\phi_{cs}$}{A score for Contrastive Search watermark.}
\nomenclature{$P$, $P_{cs}$, $P_{tp}$}{P-values indicating the likelihood of observing the test results under the null hypothesis.}
\nomenclature{$\Phi$, $F_{\chi^2}$}{Cumulative distribution functions used to calculate p-values in hypothesis testing for watermark detection.}

\clearpage
\onecolumn

\section{Additional Results}
\subsection{Rating and perplexity comparison between \algo and KGW}~\label{app:perp}
\begin{figure*}[!ht]
    \centering
    \begin{subfigure}[t]{0.45\linewidth}  
        \centering 
        \includegraphics[width=\textwidth]{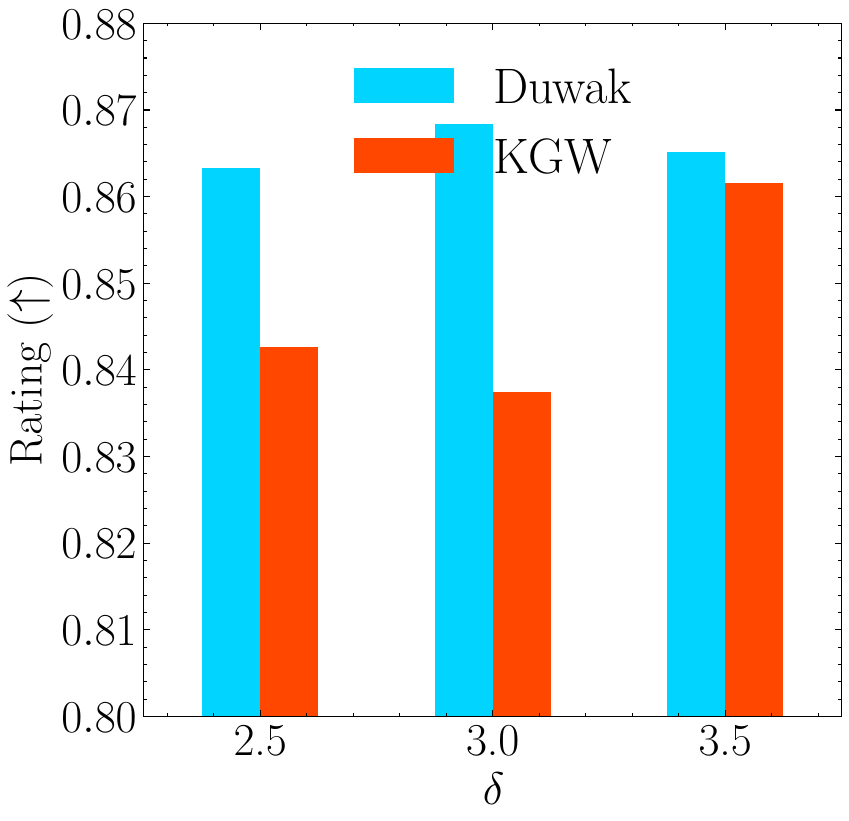} 
        \caption{No attack.}
        \label{fig:min_efficiency}
    \end{subfigure}
    ~
    \begin{subfigure}[t]{0.45\linewidth}  
        \centering 
        \includegraphics[width=\textwidth]{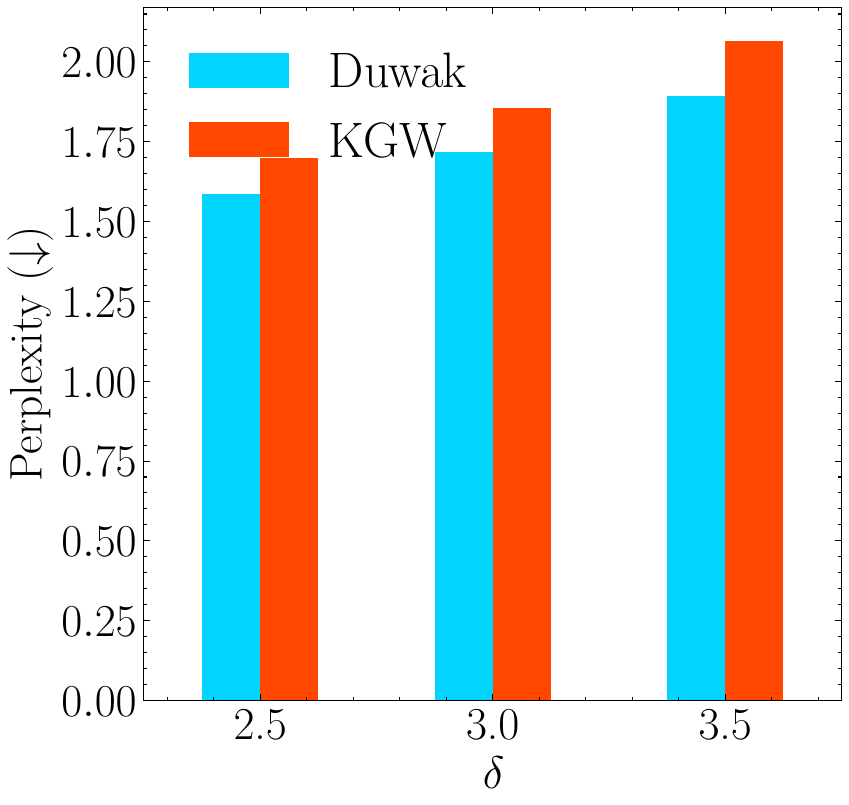} 
        \caption{Paraphrase attack.}
        \label{fig:min_efficiency_para}
    \end{subfigure}
    \caption{Comparative analysis of \algo and KGW with identical hyper-parameters under varying $\delta$, detection efficiency ($\downarrow$).
    }\label{fig:perp}
\end{figure*}
\subsection{Empirical false positive rates}

To assess the theoretical false positive rate (p-value) through empirical means, we utilize a the following methodology to identify unwatermarked text within the Wikitext dataset. For each watermark, we examine 10,000 samples, each with an average length of 260 tokens. The empirical false positive rate is determined by the proportion of texts erroneously identified as watermarked at the p-value threshold.
\begin{figure*}[!ht]
    \centering
    \begin{subfigure}[t]{0.45\linewidth}  
        \centering 
        \includegraphics[width=\textwidth]{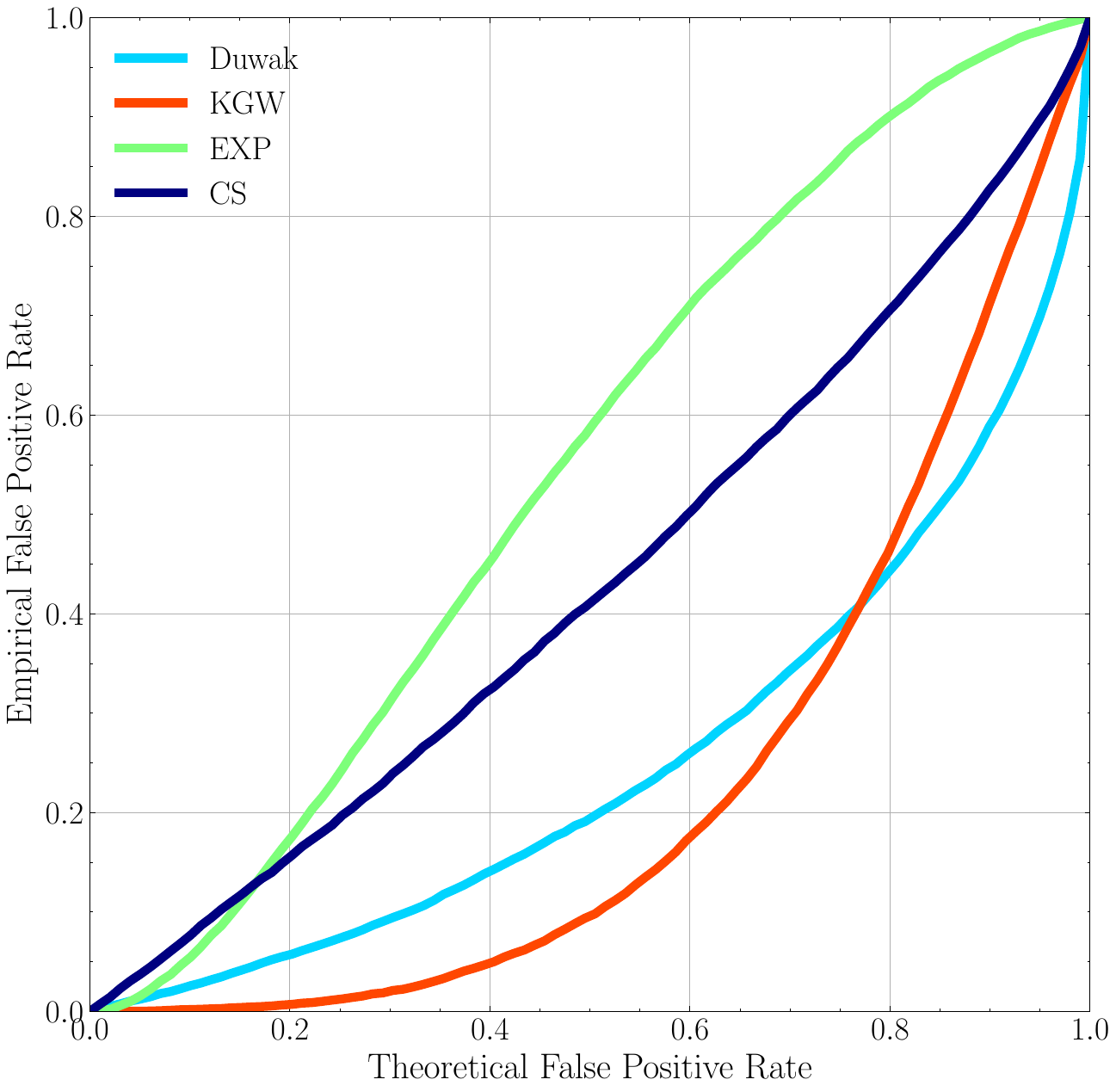} 
        \caption{Complete comparison.}
        \label{fig:fpr_1}
    \end{subfigure}
    ~
    \begin{subfigure}[t]{0.45\linewidth}  
        \centering 
        \includegraphics[width=\textwidth]{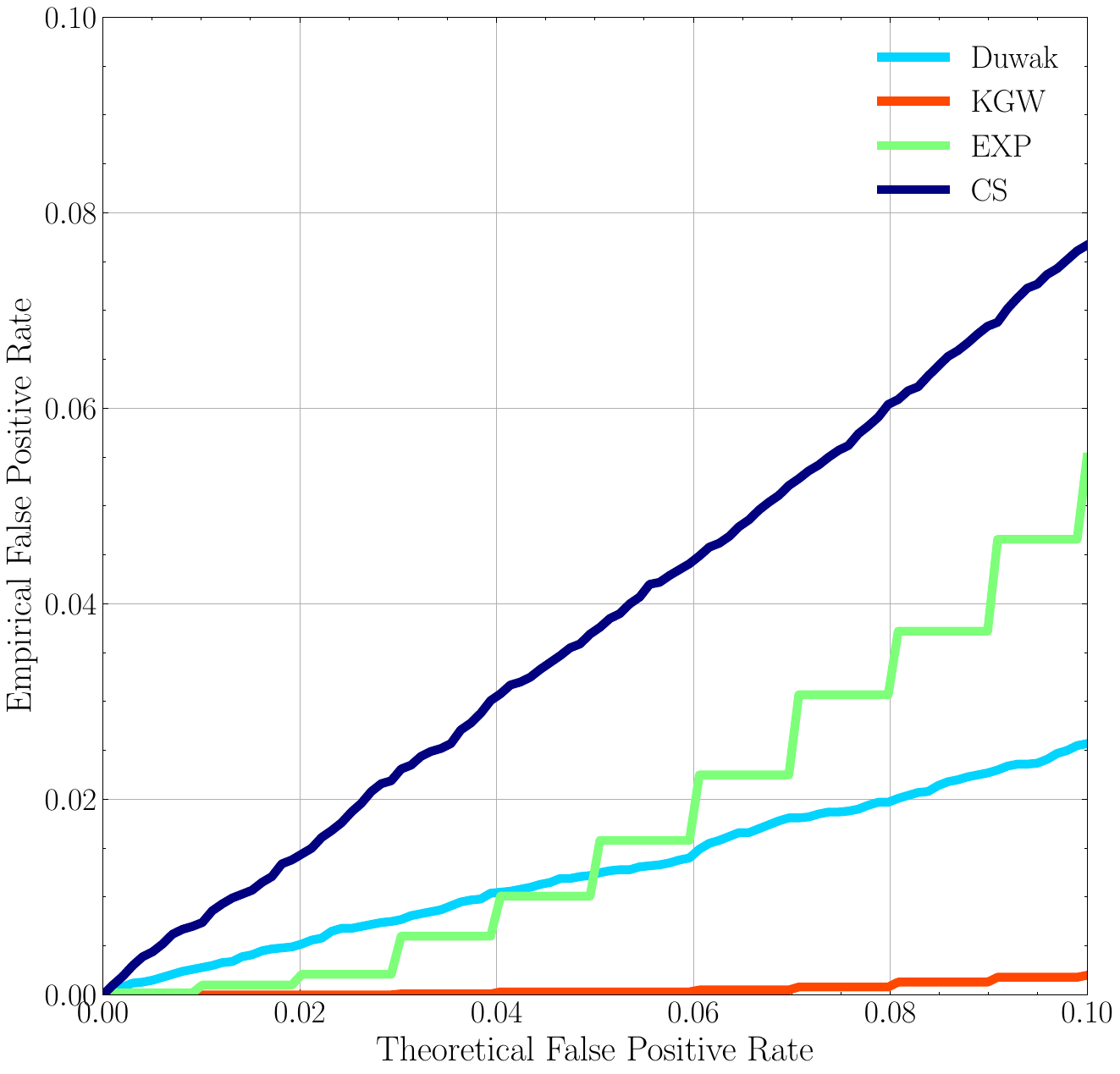} 
        \caption{Zoomed-in version of Fig.~\ref{fig:fpr_1} from $0$ to $0.1$.}
        \label{fig:fpr_0.1}
    \end{subfigure}
    \caption{Comparison of empirical false positive rate and theoretical false positive rate for different watermarks}\label{fig:fpr}
\end{figure*}

We observe that our algorithm does not result in an empirical false positive rate (FPR) higher than the theoretical FPR. Moreover, all methods tend to overestimate the false positive rate when the p-value is lower than 0.1, particularly in the case of KGW.

\subsection{Detection Efficiency Under Different p-value Thresholds}

\begin{figure*}[!ht]
    \centering

    \includegraphics[width=0.7\linewidth]{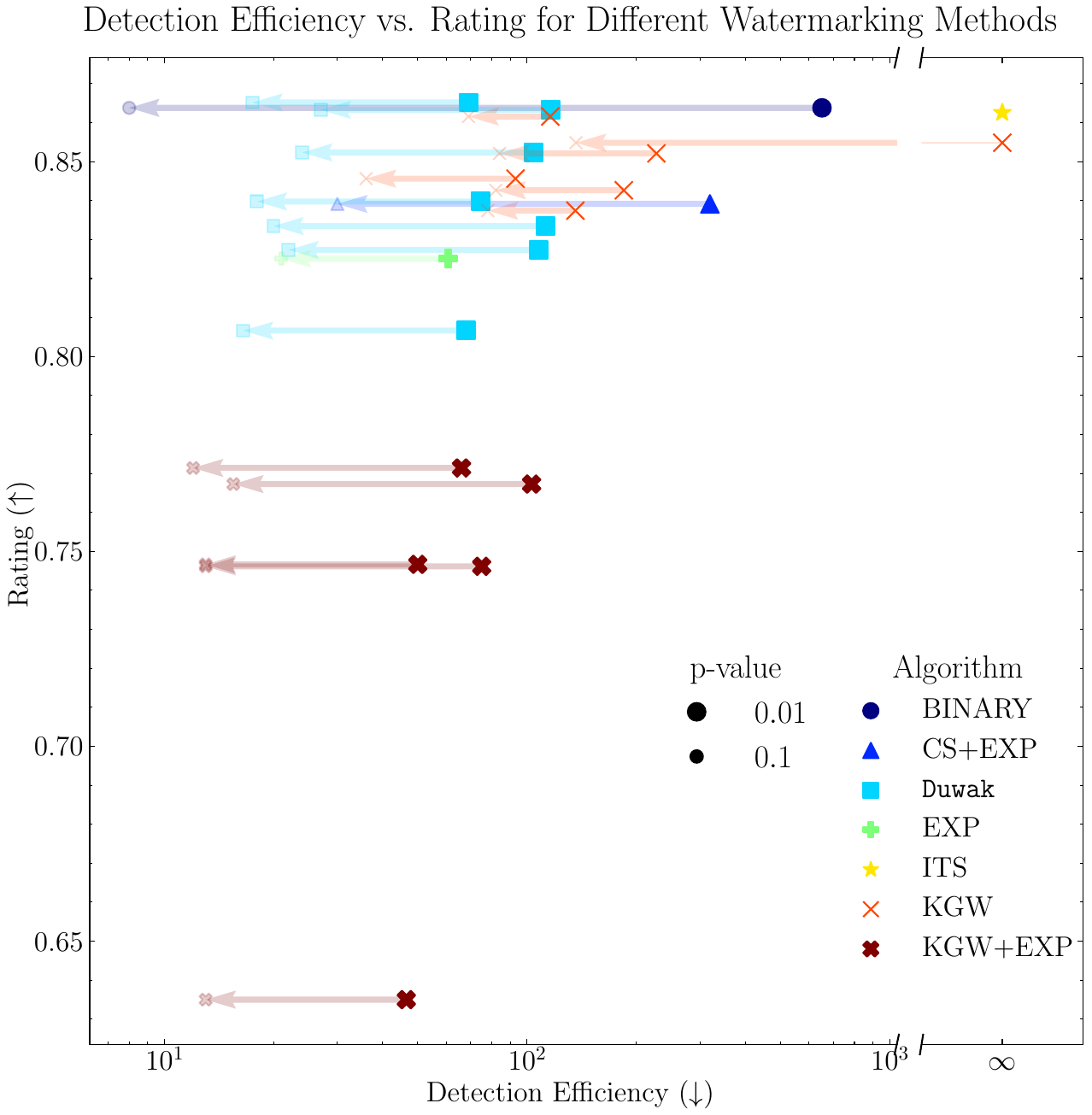}
    \caption{Detection efficiency vs. rating under different watermarking methods and hyper-parameter settings with p-value $0.01$ and $0.01$. Arrows are drawn between the corresponding configurations with different p-values to indicate the reduction of detection tokens required for a more lenient p-value.}\label{fig:p-value_detection_tokens_0.0}
\end{figure*}

\subsection{Ablation Study}
\cmr{
We conducted additional experiments on the Llama2-7b model, as shown in Table~\ref{tab:watermark_abaltion}, to assess the effectiveness of our \algo when both components are active as well as to assess the effectiveness of each individual component (KGW and CS) on the process. Table~\ref{tab:watermark_abaltion} summarizes the detection efficiency results, defined as the median number of minimal tokens needed for detection when the p-value reaches 0.02. These findings clearly demonstrate that the combined use of both the KGW watermark and the contrastive search watermark within our \algo results in superior detection efficiency compared to the performance the individual components.
}

\noindent\begin{table}[!hb]
\centering
\resizebox{0.6\linewidth}{!}{
\begin{tabular}{llccc}
\toprule
Attack & Conf. & \algo & KGW in \algo& CS in \algo \\ \midrule
None            &                & 83             & 89                    & 269                  \\ 
Contraction     &                & 81             & 87                    & 241                  \\ 
Lowercase       &                & 97             & 114                   & 233                  \\ 
Repetition\&deletion &         & 76             & 82                    & 300                  \\ 
Misspelling     & 25\%           & 84             & 91                    & 543                  \\  
                & 50\%           & 88             & 99                    & 228                  \\ 
Swap            & 5\%            & 84             & 96                    & 342                  \\  
                & 10\%           & 89             & 102                   & 206                  \\ 
Synonym         & 25\%           & 88.5           & 92                    & 247.5                \\ 
                & 50\%           & 85.5           & 104                   & 275.5                \\ 
                & 75\%           & 91             & 108                   & 259.5                \\  
                & 100\%          & 97.5           & 125                   & 351                  \\ 
Translation     & FR             & 105            & 153                   & 675                  \\  
                & RU             & 107            & 148.5                 & \textgreater1024     \\ 
TypoAttack      & 5\%            & 85             & 150                   & 517                  \\ 
                & 10\%           & 152            & 225                   & 985                  \\ 
Paraphrase      & GPT3.5         & 184            & 312                   & 1012                 \\ \bottomrule
    \end{tabular}
}
\caption{Attacked detection efficiency on Llama2-7b for \algo and it's components, lower is \textbf{better}.}\label{tab:watermark_abaltion}
\end{table}
\section{Evaluation Setup}
\subsection{Prompt}~\label{app:prompt}
For evaluation, we use open-ended generation and MarkMyWords'~\cite{piet2023mark} structured tasks. 

\emph{Open-ended text generation:} Following~\citet{DBLP:journals/tmlr/SuC23}, 40 WebText corpus token prompts are used to generate completions of up to 200 tokens.
Comparing the quality of watermarked generations to the datasets' human-completions indicates the watermarkings' relative effectiveness.

\emph{MarkMyWords generation tasks:} Additionally, we include three tasks from the MarkMyWords dataset---book reports, story generation, and fake news generation---to mirror realistic scenarios where watermarking allows for harm mitigation and accountability.

\subsection{Evaluation metrics}~\label{app:metrics}
We use the following metrics to assess the performance and effectiveness of watermark methods. 

\emph{Diversity:} Accounts variance in generated content using repetition under varying \(n\)-grams (\(n \in \{2,3,4\}\)). 

\emph{MAUVE:}~\cite{Krishna:NIPS:21:MAUVE} Measures the similarity between generated and human-written text token distributions. Higher MAUVE shows better resemblance to human text. 

\emph{Rating:} Automated evaluation with the GPT-3.5 Turbo API, rating the relevance and fluency of watermarked texts on a 100-point scale based on zero-shot prompting.

\emph{Detection efficiency:} The minimum token count required for watermark detection for a  type-I error rate (p-value).
Thus ensuring a precise balance between text length and the efficacy of watermark detection, highlighting our watermarking technique's efficiency.


\subsection{Assessment Guidelines for Rating Responses}
To systematically evaluate the text quality, each text will be graded out of a total of 100 points by GPT-3.5. The grading criteria are divided into four key categories, with points allocated as follows:

\textbf{Accuracy (20 points)}: This measures the correctness and relevance of the response in relation to the provided prompt. Points are awarded based on how well the response adheres to factual accuracy and pertinence to the initial query or statement.

\textbf{Detail (20 points)}: This assesses the comprehensiveness and depth of the response. A higher score is given for responses that cover the topic thoroughly, providing a nuanced understanding of the subject matter.

\textbf{Grammar and Typing (30 points)}: This category evaluates the grammatical and typographical precision of the response. A flawless submission, free from spelling errors, punctuation mistakes, and grammatical inaccuracies, will receive full points.

\textbf{Vocabulary (30 points)}: This criterion looks at the appropriateness and richness of the language used. Responses that utilize a diverse vocabulary appropriately and effectively to convey ideas will score higher in this category.

Points should be deducted for any deficiencies observed within each category. The total score, calculated out of 100, should be presented at the beginning of the evaluative feedback.

\textit{Prompt Template:}

\begin{verbatim}
[INST] <<SYS>> You are given a prompt and a response,
and you need to grade the response out of 100 based on: 
Accuracy (20 points) - correctness and relevance to the prompt; 
Detail (20 points) - comprehensiveness and depth; 
Grammar and Typing (30 points) - grammatical and typographical accuracy;
Vocabulary (30 points) - appropriateness and richness. 
Deduct points for shortcomings in each category. 
Give a total grade at the first line of the response. <</SYS>> 
Prompt: {}
Response: {}
[/INST] Grade out of 100: 
\end{verbatim}

This structured approach ensures a comprehensive and fair assessment of GPT-3.5's responses, facilitating a consistent evaluation process.

\clearpage
\section{Generated text}

\begin{longtable}{ccp{\linewidth-4cm}}
\caption{Completions to prompt ``[INST] <<SYS>> You are a helpful assistant. Always answer in the most accurate way. <</SYS>> Write a story about a retired astronaut that joins a risky interstellar rescue mission.. [/INST]'', limited to first 3 sentences.}\label{tab:generation_story} \\
\toprule
 &  & response \\
generator & delta &  \\
\midrule
BINARY & 1.5 &   John "Hawk" Hawkins had seen it all during his illustrious career as an astronaut. He had flown on dozens of missions to space, including the historic first manned voyage to Mars, and had even broken a record for the longest time spent in space. But after decades of flying through the cosmos, Hawk decided it was time to hang up his space suit and enjoy his well-deserved retirement.\textbackslash{}n\textbackslash{}nThat was until he received a distress call from a nearby planet that had been devastated by a catastrophic event \\
\cline{1-3}
CS & 2.5 &  Title: The Call of Duty\textbackslash{}n\textbackslash{}nRetired astronaut John "Hawk" Hawkins had seen it all during his illustrious career in space exploration. From the vastness of the cosmos to the intricate workings of the human body, he had experienced it all. But at the age of 65, he thought he had left his adventurous days behind \\
\cline{1-3}
CS+EXP & 2.5 &  Title: The Call of Duty\textbackslash{}n\textbackslash{}nJohn "Hawk" Hawkins, a retired astronaut in his early sixties, had grown accustomed to the quiet life on Earth. After decades of space travel, he had left the thrill of exploration behind and settled into a comfortable routine. However, fate had other plans.\textbackslash{}n\textbackslash{}nOne fateful evening, while watching the news, Hawk's eyes widened as he listened to the urgent message from Mission Control: an interstellar rescue mission was underway, and they needed his expertise \\
\cline{1-3}
\multirow[t]{2}{*}{CS+KGW} & 2.5 &  Title: The Call of Duty\textbackslash{}n\textbackslash{}nRetired astronaut John "Jack" Jackson had spent his entire adult life working for NASA, logging over a hundred days in space and earning a reputation as one of the best pilots in the business. But after a near-fatal accident on a mission to Mars, he was forced to retire and leave his beloved space travel behind.\textbackslash{}n\textbackslash{}nYears went by, and Jack found himself growing restless and unfulfilled in his post-NASA life. He spent his days golfing and fishing, but something was missing \\
\cline{1-3}
EXP & 1.5 &   Title: The Courageous Retiree\textbackslash{}n\textbackslash{}nJohn "Hawk" Hawkins, a grizzled astronaut in his early sixties, had long since left his spacefaring days behind. Or so he thought. When he received the phone call from Mission Control, his heart raced with a mixture of excitement and trepidation \\
\cline{1-3}
\multirow[t]{5}{*}{KGW}
 & 2.5 &   Title: The Call of Duty\textbackslash{}n\textbackslash{}nMission Briefing:\textbackslash{}n\textbackslash{}nMission Objective: Retrieve the last surviving member of the Icarus Expedition, stranded on a hostile exoplanet.\textbackslash{}n\textbackslash{}nMission Location: Planet Xylophia-IV, located 4.5 light years from Earth.\textbackslash{}n\textbackslash{}nMission Profile: The Icarus Expedition was sent to explore Xylophia-IV over 5 years ago. Upon reaching the planet's atmosphere, their spacecraft encountered unexpected storms and malfunctions, causing them to crash-land on the hostile surface. The surviving member of the expedition, Captain Emma Taylor, has been stranded on the planet for over 4 years, with limited supplies and medical equipment \\
\bottomrule
\end{longtable}

\begin{longtable}{ccb{\linewidth-4cm}}
\caption{Completions to prompt ``[INST] <<SYS>> You are a helpful assistant. Always answer in the most accurate way. <</SYS>> Write a book report about 'A Clockwork Orange', written by Anthony Burgess. [/INST]'', limited to first 3 sentences.}\label{tab:generation_summary} \\

\toprule
 &  & response \\
generator & delta &  \\
\midrule
BINARY & - &   Sure, I'd be happy to help you with a book report on "A Clockwork Orange" by Anthony Burgess!\textbackslash{}n\textbackslash{}nIntroduction:\textbackslash{}n\textbackslash{}n"A Clockwork Orange" is a dystopian novel written by Anthony Burgess in 1962. The novel is set in a future society where violence and sexuality are Conditioned into the population through a peculiar Conditioning method called the "Ludovico technique." The story follows the protagonist, Alex, a young man who is imprisoned for his vicious crimes and undergoes the Ludovico treatment to reduce his criminal urges. However, upon his release, Alex finds that he is unable to enjoy his favorite pastime – raping and Ultraviolence – due to the side effects of the treatment.\textbackslash{}n\textbackslash{}nPlot Summary:\textbackslash{}n\textbackslash{}nThe novel begins with Alex, a charismatic and violent young man, who lives in a dystopian society where Conditioning is a norm \\
\cline{1-3}
CS & - &  Title: A Clockwork Orange\textbackslash{}nAuthor: Anthony Burgess\textbackslash{}nPublished: 1962\textbackslash{}nPages: 208\textbackslash{}n\textbackslash{}nBook Report:\textbackslash{}n\textbackslash{}n"A Clockwork Orange" is a dystopian novel written by anthony Burgees in 1962. The story takes place in a future society where violence and sexuality are socially accepted and encouraged. The protagonist, Alex DeLarge, is a charismatic and intelligent young hoodlum who leads a gang of teenage thugs in a world where "ultraviolence" is the norm.\textbackslash{}n\textbackslash{}nThe novel explores themes of free will, morality, and rehabilitation through Alex's journey \\
\cline{1-3}
CS+EXP & - &  Title: A Clockwork Orange\textbackslash{}nAuthor: Anthony Burgess\textbackslash{}nPublished: 1962\textbackslash{}nPages: 208\textbackslash{}n\textbackslash{}nBook Report:\textbackslash{}n\textbackslash{}n"A ClockWORK Orange" is a dystopian novel written by Anthony Burgee in 1962. The story takes place in a future society where violence and sexual deviance are socially accepted and encouraged. The protagonist, Alex, is a charismatic and intelligent young hoodlum who narrates the events of the novels with a unique slang and dialect.\textbackslash{}n\textbackslash{}nThe novel begins with Alex and his droogs (friends) committing a series of violent crimes, including rape and murder \\
\cline{1-3}
\multirow[t]{1}{*}{CS+KGW} & 2.5 &  Sure, I'd be happy to help you with your book report on "A Clockwork Orange" by Anthony Burgess!\textbackslash{}n\textbackslash{}nTitle: "A Clockwork Orange"\textbackslash{}nAuthor: Anthoyn Burgess\textbackslash{}nPublication Date: 1962\textbackslash{}nGenre: Dystopian Novel, Satire, Young Adult\textbackslash{}n\textbackslash{}nPlot Summary:\textbackslash{}nSet in a dystopian future, "A Clockwork Orange" follows the story of Alex, a charismatic and violent young hoodlum, who is arrested and sentenced to a controversial rehabilitation program called the "Ludovico technique." This program is designed to condition Alex to detest violence and sex, and to promote a sense of empathy and compassion towards others. However, the program has unintended consequences, and Alex becomes increasingly isolated and disillusioned with society.\textbackslash{}n\textbackslash{}nThroughout the novel, Burgess explores themes of free will, morality, and the nature of evil. Burgess challenges the reader to question the morality of punishing individuals for their actions, and to consider the ethics of manipulating people's emotions and behaviors through conditioning.\textbackslash{}n\textbackslash{}nCharacters:\textbackslash{}n\textbackslash{}n* Alex: The protagonist of the novel, Alex is a violent and charismatic young hoodlum who is the narrator of the story \\
\cline{1-3}
EXP & 1.5 &   Title: A Clockwork Orange\textbackslash{}nAuthor: Anthony Burgess\textbackslash{}nPublication Date: 1962\textbackslash{}nGenre: Dystopian Novel, Horror, Science Fiction\textbackslash{}n\textbackslash{}nBook Report:\textbackslash{}n\textbackslash{}nSet in a dystopian near-future, Anthony Burgess's "A Clockwork Orange" follows the story of Alex, a charismatic young hoodlum who, along with his gang, commits gruesome acts of violence and rape. The novel explores themes of free will, morality, and the impact of societal conditioning on human behavior.\textbackslash{}n\textbackslash{}nTitle Analysis:\textbackslash{}nThe title "A Clockwork Orange" is a clever play on words, referencing both the novel's dystopian setting and the lead character's name. The term "clockwork orange" is a contradictory phrase, as oranges are typically associated with nature and growth, while clockwork implies mechanical, predetermined movement \\
\cline{1-3}
\multirow[t]{1}{*}{KGW}
 & 2.5 &   Title: A Clockwork Orange\textbackslash{}nAuthor: Anthony Burgess\textbackslash{}nPublished: 1962\textbackslash{}nPage Count: 307\textbackslash{}nGenre: Dystopian Fiction, Satire, Novel\textbackslash{}n\textbackslash{}nSummary:\textbackslash{}n\textbackslash{}nA Clockwork Orange is a dystopian novel written by Anthony Burgess, published in 1962. The story takes place in a futuristic society where violence and sexuality are condoned and encouraged. The protagonist, Alex DeLarge, is a charismatic young hoodlum who is sentenced to a form of experimental conditioning called the "Ludovico technique" in an effort to reform him \\
\bottomrule
\end{longtable}


\end{document}